\documentclass{article}
\usepackage{spconf,amsmath,graphicx}

\usepackage{tabularx}
\usepackage{adjustbox}
\usepackage{graphicx}
\usepackage{fancyvrb}

\usepackage{booktabs}
\usepackage{multirow}
\usepackage{float}
\usepackage{svg}

\usepackage{enumitem}

\usepackage{amsmath,amssymb,amsfonts}
\usepackage{graphicx}
\usepackage{textcomp}
\usepackage{xcolor}
\usepackage{booktabs} 
\usepackage{float}
\usepackage{algorithm,algorithmicx,algpseudocode}
\usepackage{multirow}
\usepackage{enumitem}
\usepackage{mathtools}
\usepackage{url}

\usepackage{enumitem}
\usepackage{tikz}

\title{Robust ADAS: Enhancing Robustness of Machine Learning-based Advanced Driver Assistance Systems for Adverse Weather}
\name{Muhammad Zaeem Shahzad, Muhammad Abdullah Hanif, and Muhammad Shafique
}
\address{eBRAIN Lab, New York University Abu Dhabi (NYUAD), UAE \\
\{ms12297, mh6117, muhammad.shafique\}@nyu.edu}
\begin{document}
%
\maketitle
\begin{abstract}
In the realm of deploying Machine Learning-based Advanced Driver Assistance Systems (ML-ADAS) into real-world scenarios, adverse weather conditions pose a significant challenge. Conventional ML models trained on clear weather data falter when faced with scenarios like extreme fog or heavy rain, potentially leading to accidents and safety hazards. This paper addresses this issue by proposing a novel approach: employing a Denoising Deep Neural Network as a preprocessing step to transform adverse weather images into clear weather images, thereby enhancing the robustness of ML-ADAS systems. The proposed method eliminates the need for retraining all subsequent Depp Neural Networks (DNN) in the ML-ADAS pipeline, thus saving computational resources and time. Moreover, it improves driver visualization, which is critical for safe navigation in adverse weather conditions. By leveraging the UNet architecture trained on an augmented KITTI dataset with synthetic adverse weather images, we develop the Weather UNet (WUNet) DNN to remove weather artifacts. Our study demonstrates substantial performance improvements in object detection with WUNet preprocessing under adverse weather conditions. Notably, in scenarios involving extreme fog, our proposed solution improves the mean Average Precision (mAP) score of the YOLOv8n from 4\% to 70\%.
\end{abstract}

\begin{figure}[h]
\centering
\includegraphics[width=\linewidth]{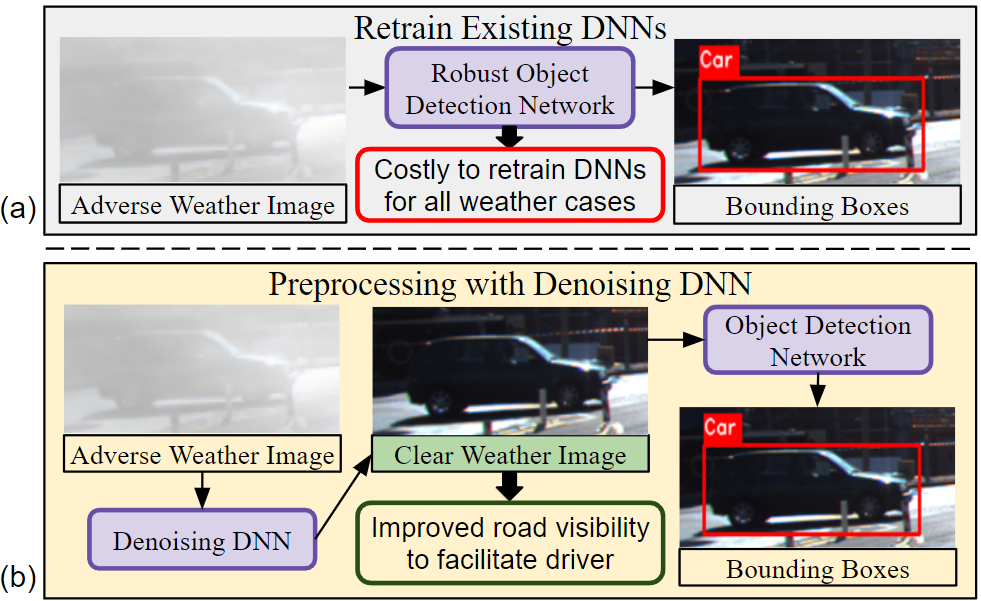}
\caption{(a) Retraining all subsequent DNNs in the ML-ADAS framework to adapt to domain shifts in weather is costly. (b) Developing a Denoising DNN as a preprocessing step before feeding input to the ML-ADAS DNNs avoids retraining and outputs clear weather images.}
\label{fig1}
\end{figure}

\begin{figure}[h]
\centering
\includegraphics[width=0.9\linewidth]{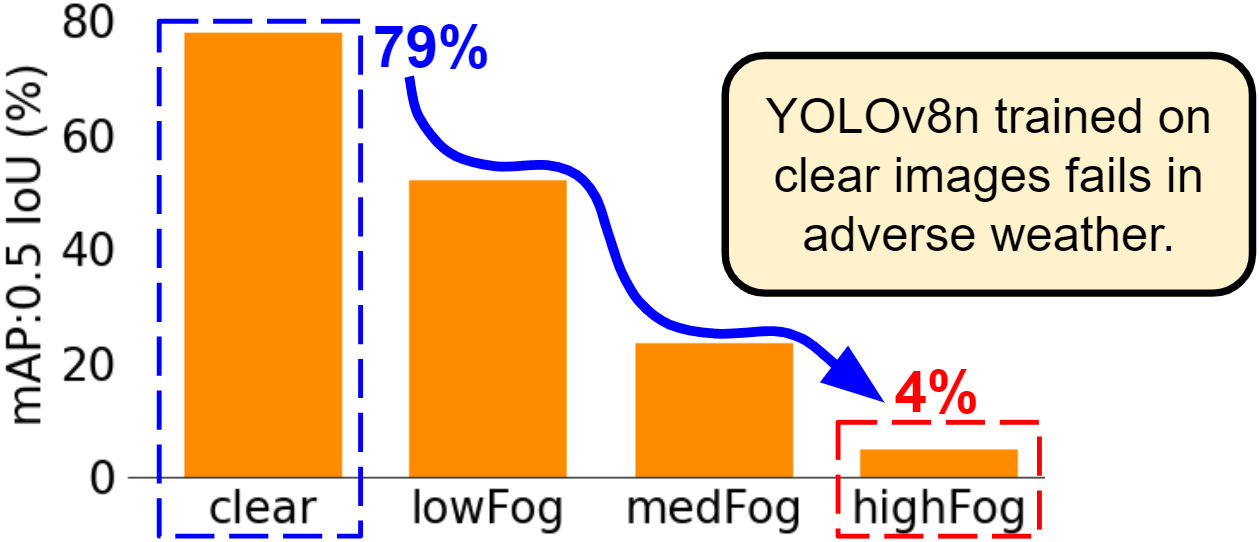}
\caption{Performance of the YOLOv8n object detector, trained on the KITTI dataset, in adverse weather conditions.}
\label{fig2}
\end{figure}


\section{Introduction}
A major challenge in deploying ML-ADAS in the real world are adverse weather conditions \textcolor{black}{~\cite{survey1, survey2}}. An object detector, such as the models in the popular YOLO \textcolor{black}{~\cite{yolov3, yolov7, ultralytics, survey3}} series, trained on clear weather images will inevitably fail to make predictions in adverse weather like extreme fog. This is because DNNs adapt to the domain presented in the training dataset and even a slight shift in the target domain will fall outside of the DNN's knowledge domain \textcolor{black}{~\cite{domainShift}}. For example, if the dataset contains only clear weather images, then the robustness of this model against adverse weather becomes an issue leading to potential accidents and safety risks. Thus, all of our DNNs need to be robustified to experience minimal performance losses in adverse weather.

An even bigger challenge is obtaining a dataset that consists of different weather conditions. Naturally, it is unsafe driving in extreme weather like rainstorms or dense fog. This is why no road scene dataset is publicly available to robustify ML-ADAS features. Some attempts have been made in the field of object detection \textcolor{black}{~\cite{wunet_prop}}. However, these datasets only record mild weather conditions where safe driving is still possible. To offer ADAS functionalities in the real world, we need to address extreme weather cases. Furthermore, a complex ML-ADAS comprises of additional functionalities like Lane Lines Detection and Distance Estimation which do not have extensive adverse weather datasets. For this reason, we decide to use multiple image augmentations to simulate adverse weather conditions and create a synthetic dataset. We shortlist fog, rain, and snow as the conditions to simulate.

Figure~\ref{fig1} highlights the strategies to address the domain shift. One approach, demonstrated in Figure~\ref{fig1}a, is to retrain all the DNNs in the system on images under different weather conditions. With this, all DNNs adapt to all weather domains. However, this process is extremely computationally expensive and time consuming. Furthermore, there is no facilitation for the driver in terms of observing the road scene ahead. While the driver may be alerted with audio cues from the robust ML-ADAS DNNs, they will still be operating blind. Thus, visualizing an enhanced road scene with all adverse weather-related noise completely removed still takes precedence in driver facilitation. 

Therefore, as described in Figure~\ref{fig1}b, instead of retraining all networks, we develop a Denoising DNN that essentially processes the input adverse weather image and provides a clear weather image. Using this module as a preprocessing step before feeding the image to all subsequent ML-ADAS DNNs retains the performance of these networks since they receive filtered images belonging to the clear weather domain. Therefore, we avoid the costly step of retraining all the subsequent DNNs in the pipeline. Most importantly, with this approach, we are able to display the clear weather images to the driver for additional visual facilitation. Additionally, it allows us to use state-of-the-art open-source models for lane segmentation, object detection and depth estimation.

Designing an ADAS for edge devices is subjected to multiple real-world constraints in terms of real-time performance, energy efficiency, and safety. Thus, our primary objective is to implement the optimal trade-off between these constraints that ensures the highest degree of safety possible. Considering this, one potential drawback of our strategy described in Figure~\ref{fig1}b is the added latency of the new DNN used for image enhancement. We address this drawback by reducing the scope of the problem: we divide the input image into crops and the WUNet processes this batch of crops instead of the entire image. This allows us to shrink the network in proportion to the reduction in input image size. The details of this scheme are described in Section II.

\subsection{Motivational Case Study}

In Figure~\ref{fig2}, we demonstrate that a highly accurate object detector trained on clear images fails to retain its accuracy when employed in adverse weather conditions. In such critical conditions, the driver relies heavily on the assistance of the ML-ADAS. However, with a 4\% mAP, there is a high risk of collisions with oncoming traffic since the system is essentially blind; it cannot detect anything. Thus, it is evident that there is a need to robustify ML-ADAS DNNs against adverse weather to guarantee driver safety after deployment in the real world. The object detector, YOLOv8n, specifications and the extended KITTI dataset for adverse weather conditions is described in Section III.


\subsection{Our Novel Contributions}
 
To robustify an ML-ADAS system against multiple adverse weather conditions without retraining all DNNs employed, we propose a preprocessing DNN to construct clear images in adverse weather. Our proposed WUNet is based on the popular UNet architecture and trained on the KITTI dataset extended with augmentations to create synthetic adverse weather images. In extreme fog, the WUNet preprocessing boosts object detection performance from 4\% to 70\% mAP over the 0.5 IoU threshold. Details on the experimental setup can be found in Evaluation and Discussion.
In summary, we make the following key contributions.
\begin{enumerate}[leftmargin=*]
    \item WUNet: we develop a preprocessing DNN to construct clear images in adverse weather, before feeding the input image to the ML-ADAS modules.
    \item We synthesize an adverse weather dataset by extending the KITTI dataset to include foggy, rainy, and snowy images. 
    \item We devise an in-depth evaluation scheme to measure a denoising algorithm's impact on object detection performance in varying weather adversities.
    \item We demonstrate potential to decrease the complexity of a UNet to decrease latency for the specific task of weather-related artifact/noise removal by processing an input image as a batch of crops instead of the whole image.
\end{enumerate}


\section{Methodology for Robustifying against Adverse Weather}

\begin{figure*}[t]
\centering
\includegraphics[width=\textwidth]{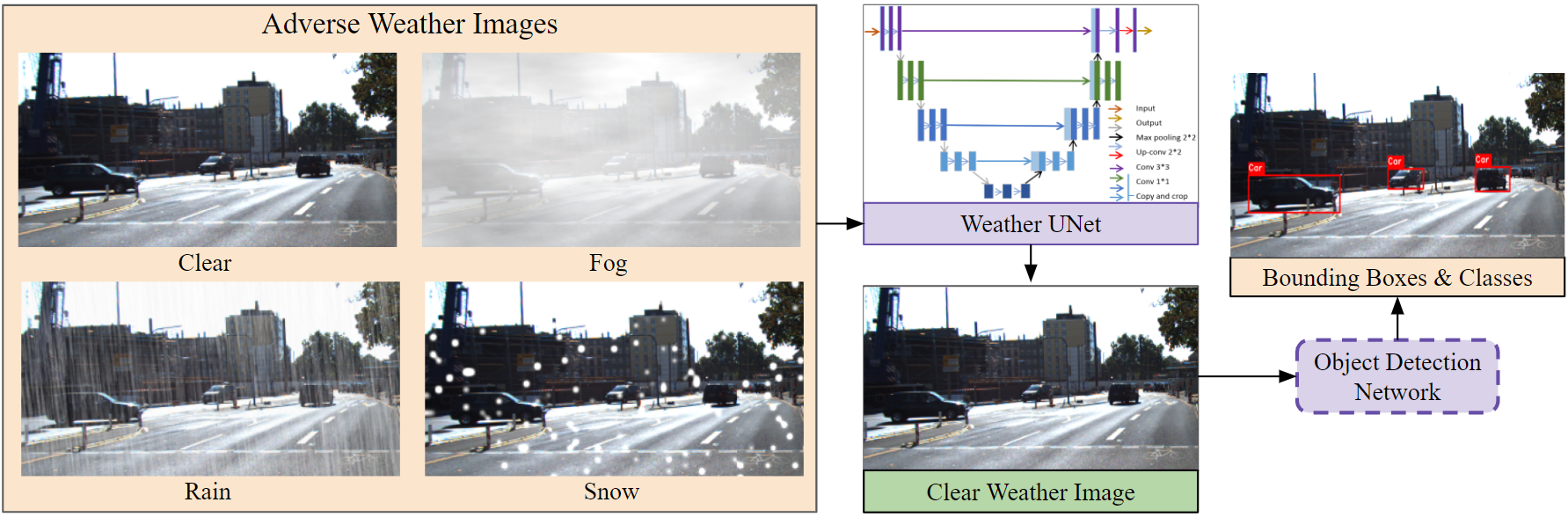}
\caption{Overview of the weather robustification methodology using the UNet for image enhancement under adverse weather conditions}
\label{fig3}
\end{figure*}

Figure~\ref{fig3} visualizes our proposed solution. We employ the popular UNet encoder-decoder architecture \textcolor{black}{~\cite{unet}} to build our Weather UNet (WUNet) to images from all weather conditions and generate a clear image. This image can then be used by the object detector and any other subsequent DNNs to avoid retraining costs. Further details are discussed in the following sections.

\subsection{Data Generation}

To address the limitation of publicly available datasets, we decide to create a synthetic dataset containing adverse weather images. With the advent of computer vision, there are multiple open-source image augmentation libraries to choose from. However, we choose the imgaug library based on its effectiveness in producing photo-realistic weather conditions \textcolor{black}{~\cite{imgaug}}. With this extensive dataset, we can train a DNN to learn to map adverse weather images to their clear counterparts. In this way, we obtain clear images that not only facilitate the driver visually, but also improve performance for all ML-ADAS features.

\subsection{Model Selection}

While similar efforts have been made to create preprocessing denoising DNNs, often the use-case is restricted to just one weather condition like fog removal \textcolor{black}{~\cite{liu2022imageadaptive}}. In contrast, we aim to address all weather cases with one DNN. The popular UNet encoder-decoder architecture performs well in similar tasks such as image super-resolution and depth estimation, where the emphasis is on image reconstruction \textcolor{black}{~\cite{superres, unet}}. Since its functionality aligns with our strategy, we adopt it in our pipeline.

\subsection{Studying Color Representations}

Training a WUNet on RGB and HSV images entails distinct approaches due to differences in color representation. RGB represents colors using three channels: red, green, and blue, which are additive. In contrast, HSV (Hue, Saturation, Value) represents colors based on perceptually relevant attributes like hue, saturation, and brightness, offering a more intuitive representation. Our exploration will involve training separate WUNet models on RGB and HSV images to discern how each representation affects performance. We plan to compare the models' ability to generate a clear image from an adverse weather image using the Mean-Squared Error (MSE) metric and comparing the prediction with the original image. Overall, we aim to gain insights into the advantages and limitations of using RGB versus HSV for image enhancement.

\begin{figure}[h]
\centering
\includegraphics[width=\linewidth]{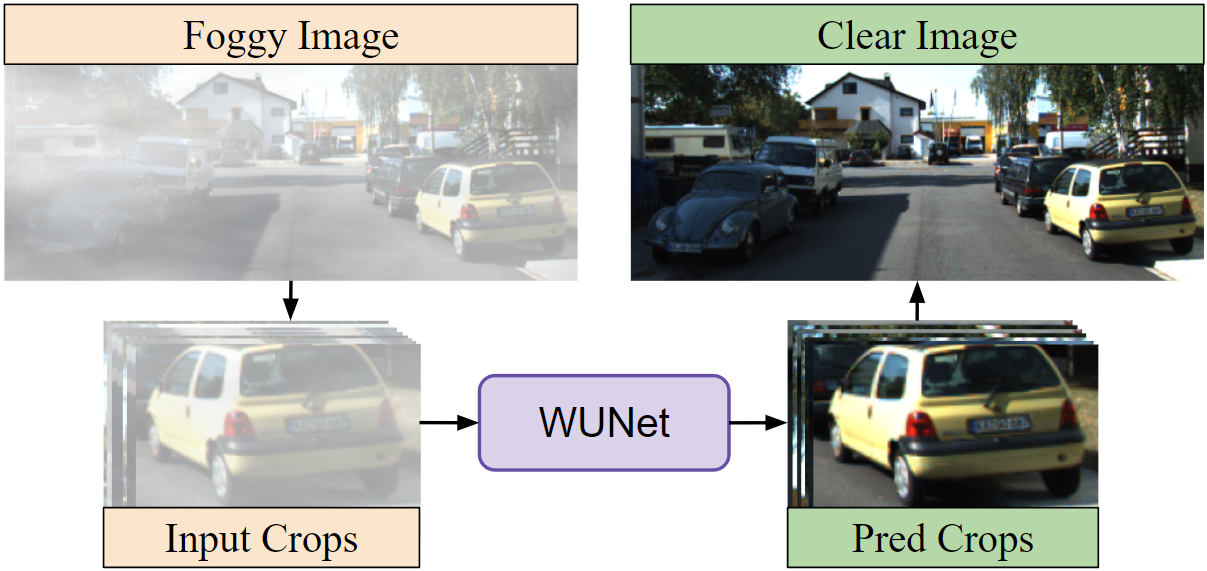}
\caption{Overview of our cost mitigation scheme. Each image is divided into crops which the WUNet processes. The prediction crops are joined together to output a full image after inference.}
\label{fig5}
\end{figure}

\begin{figure*}[t]
\centering
\includegraphics[width=\textwidth]{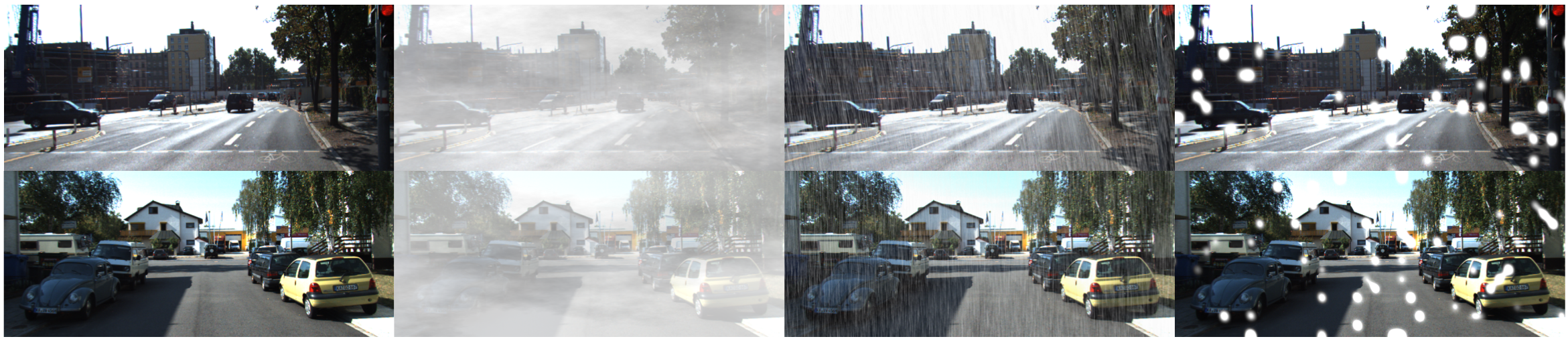}
\caption{Images from the KITTI dataset extended with image augmentations for adverse weather conditions. These images correspond to the high adversity validation sets. From left to right, in each column we show the base KITTI image, and the extreme fog, extreme rain, and extreme snow synthetic images respectively.}
\label{fig4}
\end{figure*}

\subsection{Mitigating Cost}

As mentioned earlier, the main drawback of this approach, as opposed to retraining the detector, is the additional cost of the WUNet. Figure~\ref{fig5} highlights our strategy to mitigate the additional latency introduced by the WUNet. Before implementing any DNN compression techniques like pruning and quantization \textcolor{black}{~\cite{survey4}}, we will study if the WUNet can be reduced in size with a reduction in the scope of the problem. We hypothesize that there will be no difference in WUNet performance when it processes the input image as a whole, compared with dividing it into equal-sized crops and processing this batch of crops instead. We aim to highlight that a DNN does not need to learn an extremely complex function to remove weather artifacts. Thus, with less information as input, the DNN does not need to be as large and can be reduced in size. Our study aims to highlight this potential to shrink the UNet architecture specifically in the task of noise removal. We will evaluate the difference between WUNet versions trained on either whole images or crops.

\begin{figure}[h]
\centering
\includegraphics[width=\linewidth]{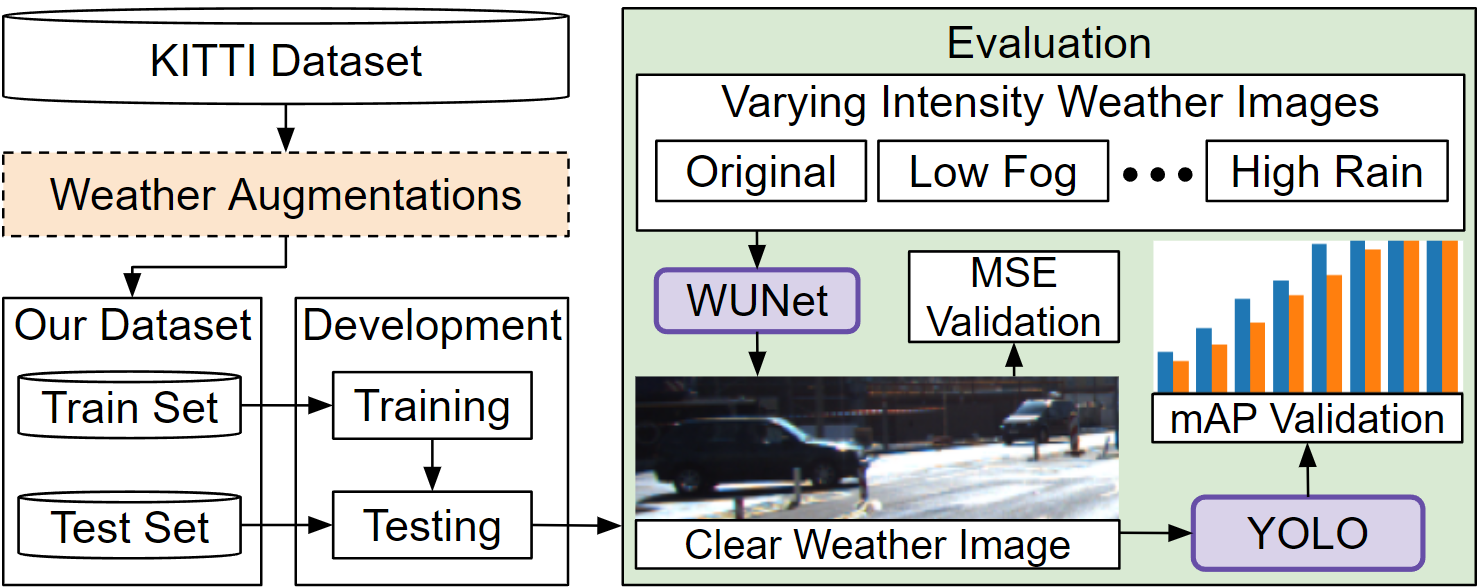}
\caption{End-to-end evaluation pipeline for WUNet combined with YOLO. The dataset used for developing the WUNet is the augmented and extended KITTI dataset}
\label{fig6}
\end{figure}

\section{Evaluation and Discussion}

Figure~\ref{fig6} describes our entire evaluation pipeline in detail. Building on the KITTI Object Detection dataset, we add synthetic adverse weather images through multiple image augmentations to create an extended dataset. Then, we use this extended dataset to train the WUNet. The evaluation phase consists of the same test set as the base KITTI dataset, but in different weather adversities. The performance of an object detector trained on the base KITTI dataset is measured on the weather validation sets with and without the WUNet preprocessing. The following sections describe this pipeline in more detail.

\subsection{Dataset}

We rely on the KITTI dataset because it consists of only clear weather images in the day time \textcolor{black}{~\cite{kitti}}. This is an important characteristic because it enables us to clearly evaluate the object detector's performance when trained on clear images and tested on adverse weather. Consequently, the impact of the WUNet on object detection becomes clear as well.

Using the train/test split provided by the authors of Dist-YOLO \textcolor{black}{~\cite{distyolo}}, we obtain 6699 images in our train, and 782 images in our test set. Next, we extend this base dataset with synthetic adverse weather images using the imgaug library \textcolor{black}{\cite{imgaug}}. We use the Fog, Rain, and Snowflakes augmentations on each image in the dataset, with the instantiation parameters for each augmentation being sampled randomly between low and extreme adversity. Along with the original, these three synthetic versions are saved with the same ground truth annotations as the original image. Thus, our dataset now consists of 26796 images in the train, and 3128 images in the test set encompassing varying intensities of each adverse weather condition. While training the WUNet, a sample is defined as the augmented image (fog, rain, snow, or original) and its ground truth is the original image. For training the object detector, the sample is defined as the image and its ground truth are the corresponding bounding box annotations in the KITTI dataset.

Furthermore, we create a total of 10 validation sets, consisting of 782 images each, by tuning the hyperparameters in the augmentations. These sets are the normal set which is the unaltered KITTI test set, and low, medium, and high intensity variations of each of the three weather augmentations. Figure~\ref{fig4} displays samples from our high adversity validation sets. With this, we can conduct an in-depth evaluation of the WUNet's weather artifact removal performance in different intensities of different weather conditions.

\begin{figure}[h]
\centering
\includegraphics[width=\linewidth]{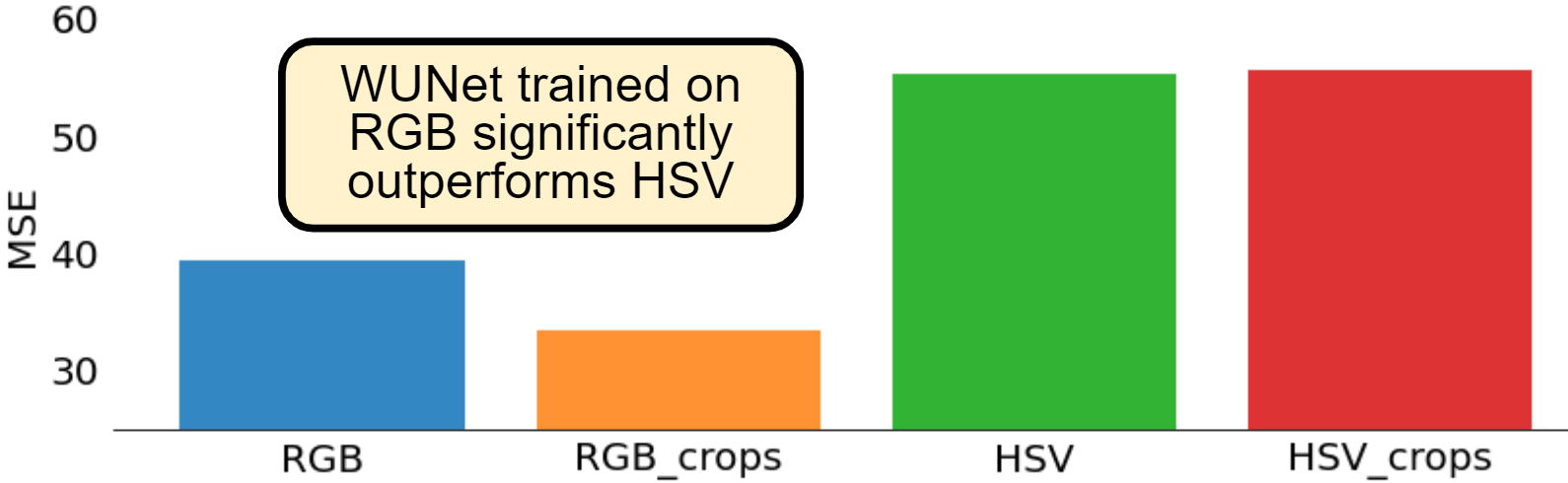}
\caption{MSE error comparison of each WUNet trained}
\label{fig7}
\end{figure}

\begin{figure*}[t]
\centering
\includegraphics[width=\linewidth]{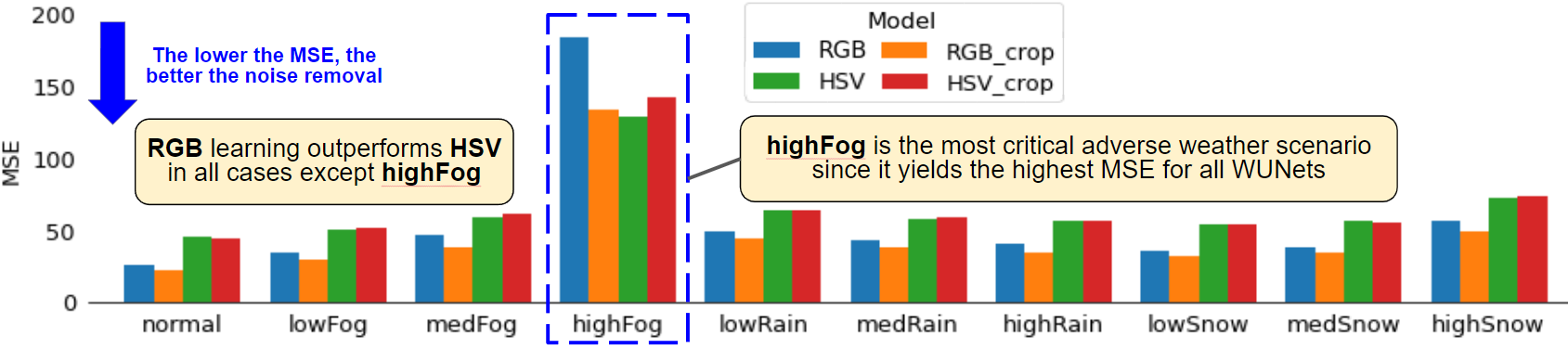}
\caption{Performance evaluation of each WUNet in terms of \textit{Mean Squared Error} (MSE).}
\label{fig8}
\end{figure*}

\begin{figure*}[t]
\centering
\includegraphics[width=\linewidth]{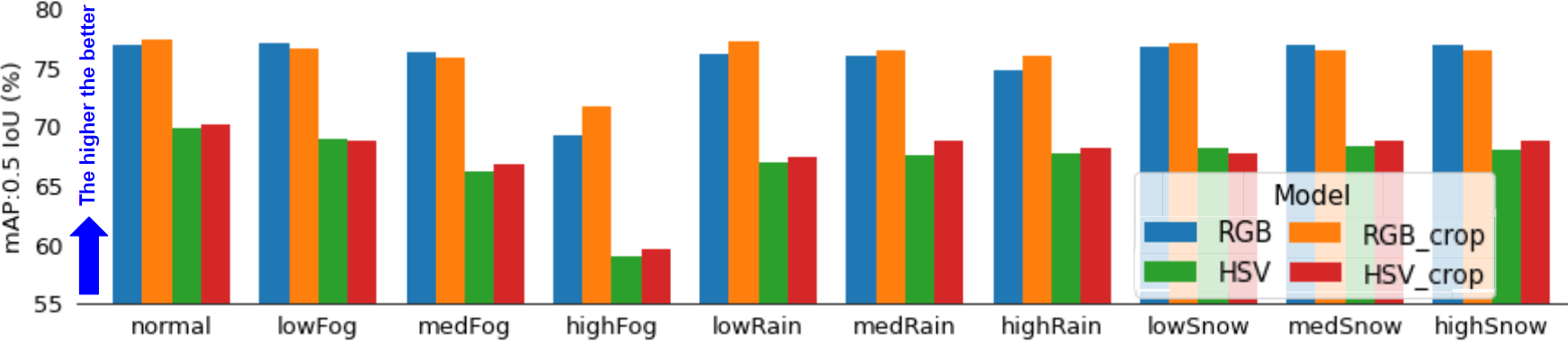}
\caption{Performance evaluation of YOLOv8n, in terms of \textit{Mean Average Precision} (mAP), with preprocessing from each WUNet trained.}
\label{fig9}
\end{figure*}

\begin{figure*}[t]
\centering
\includegraphics[width=\linewidth]{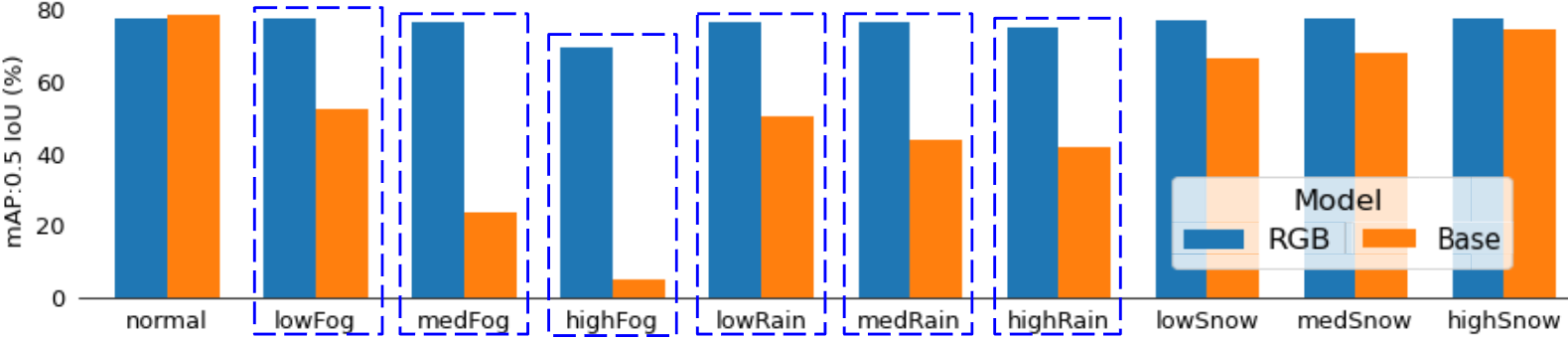}
\caption{Performance evaluation of YOLOv8n with (RGB) and without (Base) the RGB-WUNet preprocessing on adverse weather images. The dotted blue boxes highlight significant performance improvements.}
\label{fig10}
\end{figure*}

\subsection{Experimental Setup}

All experiments were conducted on the NVIDIA GeForce RTX 4090 GPU. The WUNet was trained on the extended and augmented KITTI dataset using the Adam optimizer for 200 epochs with a batch size of 24, learning rate of 0.01, and input image size of (640,200) pixels. Note that the WUNets trained on crops have input image size of (160,100) and a total of 214,368 train and 25,024 test images. This is because we arbitrarily divide each image in the augmented dataset into 8 crops, with each crop being a sample instead of the whole image. In this case, we use batch size of 160. In all experiments, we use the mean-squared loss to supervise training. The WUNet that performs the best on the test set is chosen to be evaluated end-to-end with the object detector. 

For object detection, we employ the YOLOv8 variant of size nano (n), small (s), and medium (m) from the open-sourced Ultralytics framework \textcolor{black}{\cite{ultralytics}}. These detectors have been trained over the base KITTI dataset with the same training settings as the base WUNet except for a 0.001 learning rate. We present the performance evaluation results of each of these variants in Table 1.

\begin{table}[]
\centering
\caption{Accuracy and parametric complexity of the YOLOv8 Object Detectors trained on the base (clear images) KITTI dataset.}
\resizebox{\columnwidth}{!}{%
\begin{tabular}{|l|l|l|ll}
\cline{1-3}
YOLOv8 variant & mAP over 0.5 IoU (\%) & Parameters (millions)  \\ \cline{1-3}
n              & 79.97               & 3.2                    \\ \cline{1-3}
s              & 83.46               & 11.2                   \\ \cline{1-3}
m              & 83.80               & 25.9                   \\ \cline{1-3}
\end{tabular}%
}
\end{table}

Since YOLOv8n has the lowest mAP score, it has the greatest sensitivity to weather artifacts. Thus, we use this variant in our evaluation of the WUNet end-to-end to examine the artifact removal performance in the worst case of highest detector sensitivity.

\subsection{End-to-End Evaluation with YOLOv8n}

Figure~\ref{fig7} summarizes the MSE evaluation results we obtain with training each WUNet variant. With MSE validation, we aim to measure the WUNet's ability to generate high quality clear images. The lower the MSE, the greater the WUNet's ability in removing weather artifacts. We observe that the RGB learned WUNets outperform the HSV variants by at least 40\%. Furthermore, within the RGB WUNets, training over crops yields 15\% lower MSE than training over the whole image. This is most likely because of a larger dataset in the case of crops, but we establish that processing via crops is a valid scheme. Figure~\ref{fig8} provides a closer inspection where we observe MSE error in our 10 varying adversity sets. Most notably, we observe that the most challenging and critical case to address is high fog. Next, we aim to employ the YOLOv8n's mAP performance as a metric to further determine the quality of the clear images produced.

We compare the performance of YOLOv8n, with and without each WUNet variant's preprocessing, over all the 10 cases. Note that YOLOv8n was trained only on the base KITTI dataset that consists of clear images. The validation set that corresponds to this is the normal set and we expect the highest performance here (Table 1). Figure~\ref{fig9} summarizes our findings. Averaged over all cases, the RGB variants achieve 76\% mAP compared to the HSVs' 67\%. Again, we observe that RGB-learning outperforms HSV-learning, with at least a 9\% increase in mAP in all categories. Most importantly, we have verified that the WUNet can process images as crops and retain the same performance as processing the entire image. This allows us to continue our work to shrink the size of the WUNet to increase network efficiency. 

Lastly, Figure~\ref{fig10} highlights the improvements in YOLOv8n's accuracy by using the simple RGB WUNet in the pipeline. Our results verify that the YOLOv8n trained on clear images is needs to be robustified against cases like extreme fog. Inference on the clear image provided by the WUNet compared to the extremely foggy image improves mAP score from 4\% to 70\%, a 1650\% improvement. Overall, the most significant performance boosts are seen in all fog and rain cases especially.

\subsection{Future Work}

Building upon this project, we will evaluate the WUNet on a larger dataset like the Pascal VOC dataset \textcolor{black}{\cite{voc}}. This will allow us to compare our results to the current state-of-the-art works like the Image-Adaptive YOLO that use larger datasets, as opposed to the KITTI dataset \textcolor{black}{\cite{liu2022imageadaptive}}. Additionally, research needs to be conducted to assess the DNN's performance in terms of its efficiency and additional latency impact on overall system. We have already established that the WUNet is able to process images as a batch of crops which allows us to shrink the network's size by a factor that is directly proportional to the size of the crop relative to the original image size. This is because removing weather related artifacts from the image does not require learning a very complex function. Thus, we will tune the parameters of network size and image-crop size specifically.

\section{Conclusion}

In conclusion, our study addresses the critical challenge of deploying ML-ADAS in adverse weather conditions. We demonstrate that ML-ADAS systems trained solely on clear weather data are ill-equipped to handle adverse conditions such as extreme fog or heavy rain, posing significant safety risks. To mitigate these risks, we propose a novel approach leveraging a Denoising DNN as a preprocessing step to generate clear weather images from adverse weather inputs, without the need for retraining all subsequent DNNs in the ML-ADAS pipeline. This strategy not only avoids the computational cost and time-consuming process of retraining, but also enhances driver visualization, crucial for safe navigation. 

Through extensive experimentation and evaluation, we demonstrate the efficacy of our approach using the UNet architecture, trained on an extended KITTI dataset augmented with synthetic adverse weather images. Our findings indicate significant performance improvements in object detection under adverse weather conditions, with the WUNet preprocessing leading to a substantial increase in mAP scores, particularly in scenarios involving extreme fog where we boost a 4\% mAP to 70\%. 

In light of our findings, we propose future research directions, including evaluation on larger datasets and further optimization of network parameters, i.e. input image size and size of the network specifically, for enhanced efficiency and performance. Overall, our study offers valuable insights and a promising solution to enhance the robustness and safety of ML-ADAS systems in adverse weather conditions, paving the way for safer autonomous driving experiences in challenging environments.

\subsection*{Acknowledgements}
This work was partially supported by the NYUAD Center for Artificial Intelligence and Robotics (CAIR), funded by Tamkeen under the NYUAD Research Institute Award CG010, and the NYUAD Center for Interacting Urban Networks (CITIES), funded by Tamkeen under the NYUAD Research Institute Award CG001.

\bibliographystyle{IEEEbib}
\bibliography{main.bib}

\end{document}